\documentclass[cameraready]{Interspeech}
\title{Layer-wise Probing of wav2vec 2.0 and Whisper for Consonant Cluster Reduction in African American English}

\author[affiliation={1},orcid=0009-0005-2502-1240]{Hamid}{Mojarad}
\author[affiliation={1,2},orcid=0000-0001-7382-9344]{Kevin}{Tang}

\address{
    $^1$ Department of English Language and Linguistics, Institute of English and American Studies, Faculty of Arts and Humanities, Heinrich Heine University Düsseldorf, Germany \\
    $^2$ Department of Linguistics, University of Florida, United States of America
}

\email{hamid.mojarad@hhu.de, kevin.tang@hhu.de}


\usepackage{comment}
\usepackage{array}   

\begin{document}

\maketitle

% the abstract here must exactly match the abstract entered into the paper submission system
\begin{abstract}
Self-supervised and supervised speech models are increasingly used to investigate which linguistic information their internal representations encode, and at what level of abstraction they encode it. One underexplored phenomenon is consonant cluster reduction (CCR) in African American English (AAE), a widespread phonological process and a source of automatic speech recognition (ASR) disparity. To examine how CCR is represented, we conduct speaker-independent layer-wise probing of \textit{wav2vec2-base} and \textit{Whisper-small} using two tasks: segmental reduction detection and segmental restoration of underlying cluster identity. Both models distinguish reduced and canonical forms with high accuracy. Crucially, reduced segments retain cues to their underlying stops, indicating that CCR is encoded as structured gradient phonological variation rather than simple segmental deletion. These results demonstrate structured phonological encoding of AAE CCR patterns in modern speech models.

    % 1000 characters. ASCII characters only. No citations.
    % Manuscripts submitted to Interspeech 2026 must use this document as both an instruction set and as a template. Do not use a past paper as a template. Always start from a fresh copy, and read it all before replacing the content with your own. The main changes with respect to previous years' instructions are \blue{highlighted in blue}.

    % Before submitting, check that your manuscript conforms to this template. If it does not, it may be rejected. Do not be tempted to adjust the format! Instead, edit your content to fit the allowed space. \blue{The maximum number of manuscript pages is 6 for regular papers and 10 for long papers. For regular papers, pages 5 and 6, and for long papers, pages 9 and 10, are reserved exclusively for acknowledgments, disclosures of the use of generative AI tools, and references, which may begin on an earlier page if space permits.}

    % The abstract is limited to 1000 characters. The one in your manuscript and the one entered in the submission form must be identical. \blue{In the submission form, no \LaTeX{} code is allowed.} Do not use citations in the abstract: the abstract booklet will not include a bibliography. Index terms appear immediately below the abstract.
\end{abstract}

\section{Introduction}

Modern ASR systems exhibit significant performance disparities across demographic groups, with particularly high error rates for speakers of AAE, a rule-governed variety of English shaped by historical, social, and cultural factors \cite{Mengesha_2021}. Multiple studies have documented racial bias in commercial ASR technologies, showing word error rates (WER) up to twice as high for AAE speakers compared to non-AAE speakers, even when controlling for age, gender, and content \cite{koenecke2020racial, martin20_interspeech, Martin2023}. These disparities stem in part from training data imbalances favoring Mainstream American English and from insufficient representation of AAE phonological and morphosyntactic features \cite{wassink2022uneven, Martin2023}.

\subsection{Under the hood of ASR}
To address ASR bias effectively, it is essential to move beyond purely error-based metrics like WER and examine how self-supervised and supervised speech models internally encode dialectal phonological variation. Researchers have increasingly turned to interpretability methods, particularly linear probing of hidden layer representations, to uncover what occurs under the hood of these encoders \cite{parra-2025-interpretable, Pasad_2021}.

Self-supervised models such as wav2vec 2.0 \cite{baevski2020wav2vec2} learn hierarchical speech representations from large amounts of unlabeled audio using a convolutional feature encoder followed by a Transformer encoder, trained with a contrastive objective over jointly learned quantized latent targets. This setup enables learning from unlabeled audio alone, eliminating the need for transcriptions.
In contrast, supervised models such as Whisper \cite{radford2023robust} employ a fully supervised encoder-decoder Transformer architecture trained on approximately 680k hours of labeled audio-text data for direct end-to-end transcription. Probing and layer-wise analyses have addressed diverse linguistic aspects, including acoustic-phonetic properties \cite{Pasad2023}, phonetic categorization \cite{cormac-english-etal-2022-domain}, and accent/prosodic variation \cite{yang23v_interspeech}, alongside syllable structure \cite{a-shams-etal-2024-uncovering} and phonemic restoration \cite{Shams2025} in representations learned by wav2vec 2.0 and Whisper, yet dialect-specific phonological processes such as CCR in AAE remain largely unexplored.

\subsection{Consonant cluster reduction}
%CCR in English is a systematic phonological process in which word-final clusters are simplified in patterned ways \cite{Wolfram_2017}. While reduction is shaped by both the following phonological context and properties of the cluster itself, dialects differ in how these constraints are weighted \cite{Wolfram_2017}. In some varieties, such as Standard English and Appalachian English, the following consonant plays a stronger role than cluster type, whereas in others, including Southern White working-class and Southern African American working-class speech, cluster-internal properties are the more influential factor \cite{Wolfram_2017}. This cross-dialectal variation highlights CCR as a structured but dialect-sensitive process governed by interacting phonological constraints \cite{Wolfram_2017}.

CCR in English is generally treated as a cross-dialectal feature rather than a phenomenon unique to any single community \cite{Labov_1972, Schreier_2005}. It is a systematic phonological process in which word-final clusters are simplified in patterned ways \cite{Wolfram_2017}. While reduction is shaped by both the following phonological context and properties of the cluster itself, dialects differ in how these constraints are weighted \cite{Wolfram_2017, Guy_1991, Schreier_2005}. This cross-dialectal variation highlights CCR as a structured but dialect-sensitive process governed by interacting phonological constraints \cite{Wolfram_2017}.

CCR in AAE typically involves omission of the final stop in two-consonant clusters (e.g., \textit{test} /\textipa{tEst}/ → [\textipa{tEs}]) or the penultimate consonant in a cluster of three (e.g., \textit{fists} /\textipa{fIsts}/ → [\textipa{fIs:}]) \cite{Erik_Baily_2015}. Previous work has shown that ethnicity-related dialect features, including CCR, lead to uneven ASR success across racial groups. For instance, Wassink et al. \cite{wassink2022uneven} evaluated a commercial ASR system on a multi-ethnic sample from the American Pacific Northwest (including AAE speakers) and found systematically higher phonetic error rates for nonwhite speakers, with dialectal phonological variation, including CCR, contributing to differential performance and highlighting racial bias in system output. Similarly, recent work on wav2vec 2.0 \cite{mojarad25_interspeech} confirmed small but significant WER increases from CCR that underscores its role in ASR disparity against AAE.

\subsection{Present study}
Building upon prior behavioral evidence of AAE-related bias in wav2vec 2.0 \cite{mojarad25_interspeech}, the present study aims to uncover the root of this bias in CCR by probing how it is internally encoded in \textit{wav2vec 2.0} and \textit{Whisper}. Specifically, we conduct a two-fold probing investigation to assess whether the models treat CCR as a simple segmental deletion task across layers, or whether they represent reduced realizations as closer to their canonical counterparts, in order to determine whether these internal representations can reveal the mechanisms underlying the observed AAE-related bias. We focus on frequent two-consonant clusters (e.g., /\textipa{nt}/, /\textipa{nd}/, /\textipa{st}/), %which are easier to produce articulatorily \cite{Bayley_Villarreal_2019}. We 
and perform two domain-informed probes \cite{cormac-english-etal-2022-domain} on frozen encoder representations:
\begin{itemize}
    \item \textbf{Segmental reduction detection}: We test whether encoder representations distinguish reduced and canonical cluster pronunciations, thereby assessing whether the presence or absence of a final stop is explicitly encoded across model layers. This probe directly evaluates the phonetic sensitivity of each model to CCR.
    \item \textbf{Segmental restoration}: We examine whether reduced forms, such as nasal-only realizations of underlying /\textipa{nt}/ or /\textipa{nd}/ clusters sharing an initial nasal consonant, still carry subtle cues to the dropped final stop. This allows us to test whether the original identity of the deleted segment is internally reconstructed by speech encoders.
\end{itemize}
Data and code are available on \url{osf.io}\footnote{\url{https://doi.org/10.17605/OSF.IO/FE2D7}}.

\section{Related Work}

Probing speech encoders has emerged as the standard method for dissecting the linguistic hierarchy of speech understanding \cite{Pasad2023}. For self-supervised models such as wav2vec 2.0, Pasad et al. \cite{Pasad_2021} perform a detailed layer-wise analysis and show that early transformer layers are dominated by low-level acoustic cues, mid-layers maximize phonetic and phonological information, and higher layers increasingly reflect lexical and semantic structure. Complementary work \cite{cormac-english-etal-2022-domain, kim24l_interspeech} applied phonetic and articulatory classification probes to wav2vec 2.0, showing that segmental categories and features (e.g., nasality, place) are most robustly encoded in intermediate layers. These findings support a hierarchical transformation from raw acoustics to increasingly abstract phonetic and lexical representations.

More recently, similar probing approaches have been applied to Whisper’s supervised encoder. Studies on pathological and accented speech show that Whisper’s mid-level encoder layers are particularly informative for phonetic and phonologically structured deviations from canonical speech. Batra et al.\ \cite{Batra_2025} report that different types of stuttered disfluencies are best discriminated from fluent speech using mid-to-late Whisper layers, while Yue et al.\ \cite{Yue_etal_2026} find that layers 13-15 of Whisper-medium yield peak performance for dysarthric speech detection and severity assessment. 

Closer to our research, Gessinger et al. \cite{Shams2025} investigate phonemic restoration in wav2vec 2.0 and Whisper. Their study introduces controlled perturbations, including noise overlays, noisy gaps, or silent gaps, into English words and pseudowords, then probes the models' transformer encoder layers for reconstruction of articulatory features (place, manner, voicing). They test wav2vec 2.0 and Whisper on these degraded stimuli, which mimic real-world noise or interruptions. Linear probes across layers reveal wav2vec 2.0's superior recovery, especially for words over pseudowords via lexical context; noisy gaps prove most disruptive, followed by silent gaps. This design parallels our planned CCR analysis in AAE, where we expect natural deletions (e.g., /st/ → /s/) to create analogous ``gaps'' that test the models' ability to reconstruct canonical forms from context.

Despite prior work probing wav2vec 2.0 and Whisper for various related linguistic aspects, no probing study has specifically investigated CCR in AAE, a dialectal process in which surface deletions are systematically predicted to recover canonical phonology via lexical context. This phenomenon allows us to test whether wav2vec 2.0 and Whisper internally restore deleted segments akin to human listeners, or simply encode surface realizations. Our dual-probe approach, \emph{segmental reduction detection} and \emph{segmental restoration}, applied to natural AAE data provides the first computational analysis of this widespread dialectal pattern.

\section{Methodology}
\subsection{Data Preparation}
\subsubsection{Corpus}

The Corpus of Regional African American Language (CORAAL) \cite{farrington_corpus_2021} serves as the foundational dataset for this study.
%, offering a comprehensive documentation of regional African American Language (AAL) varieties. 
The corpus provides rich linguistic resources, including audio recordings with time-aligned orthographic transcriptions in TextGrid format, featuring speaker-specific tiers at both utterance and word/phone alignment levels.

For this research, we utilized three subcorpora - DCA, DCB, and DTA - comprising a total of 156 speakers (80 men and 76 women). These subcorpora represent speakers from two distinct geographic regions, Washington, DC and Detroit, thereby incorporating regional variation into our dataset. In addition, each subcorpus includes speakers across four age groups and three socioeconomic classes. This design ensures diversity in terms of geography, gender, age, and social class, providing a broad and socially representative sample for examining CCR and supporting greater generalizability of the results.

\subsubsection{Feature extraction}

To extract features related to CCR, we employed forced alignment, following the general approach of Kendall et al.~\cite{Kendall_ing_2021}. In their study, human annotations of the sociolinguistic variable (ING) were compared against forced alignment and classifiers trained by machine learning libraries, demonstrating that automated coding approaches can approximate human performance in categorizing ING variation. Building on this methodology, we adopted a pipeline based on forced alignment to automate feature extraction in our analysis.

We used the Montreal Forced Aligner (MFA, version 2.2.17; \cite{mcauliffe17_interspeech}) together with the Carnegie Mellon University (CMU) Pronouncing Dictionary. Words missing from the CMU dictionary were first identified, after which we trained a grapheme-to-phoneme (G2P) model on the CMU lexicon and generated pronunciations for these items. The resulting pronunciations were manually inspected and added to the dictionary. For words susceptible to CCR, we generated reduced pronunciation variants based on their canonical forms in the CMU dictionary; for example, the word \textit{test} /\textipa{tEst}/ has a reduced form [\textipa{tEs}] with the final /t/ deleted. Using MFA’s \texttt{train} command, we trained a custom acoustic model on the full audio datasets. The complete corpus was then force-aligned using this trained model in combination with the expanded CMU pronunciation dictionary, and the resulting alignments formed the basis for classifying words as canonical or reduced throughout the study.

\subsubsection{Selection strategy}

MFA alignment across CORAAL's three subcorpora yielded over 85,000 tokens from CCR-prone two-consonant cluster words, distributed across 48 cluster types. Due to token scarcity in most types, we selected 12 high-frequency clusters, ensuring an internal balance between reduced and canonical realizations: /\textipa{st}/, /\textipa{nd}/, /\textipa{md}/, /\textipa{nt}/, /\textipa{sk}/, /\textipa{mp}/, /\textipa{ft}/, /\textipa{St}/, /\textteshlig\textipa{t}/, /\textipa{pt}/, /\textipa{vd}/, /\textipa{zd}/. We excluded /\textipa{l}/- and /\textipa{r}/- initial clusters (e.g., /\textipa{ld}/, /\textipa{lt}/, /\textipa{rd}/, /\textipa{rt}/), because they are subject to additional phonological processes beyond CCR. Specifically, post-vocalic /l/ may undergo L-vocalization (e.g., \textit{cold} /\textipa{koUld}/ → [\textipa{koUwd}]), and when CCR also applies, this yields [\textipa{koUw}]. Post-vocalic /r/ may undergo R-deletion (e.g., \textit{cart} /\textipa{ka:rt}/ → [\textipa{ka:t}]), and with CCR, [\textipa{ka:}]. These processes confound detection of pure CCR effects.

Following the theoretical framework of Thomas and Bailey \cite{Erik_Baily_2015}, we further refined the dataset by restricting the analysis to monomorphemic consonant clusters, excluding bimorphemic past-tense forms (e.g., stunned /\textipa{stVnd}/, bussed /\textipa{bVst}/). This restriction ensures that observed reduction patterns reflect phonological preferences rather than morphological conditioning. The resulting probing dataset comprises seven cluster types 
(/\textipa{ft}/, /\textipa{nd}/, /\textipa{nt}/, /\textipa{st}/, 
/\textipa{sk}/, /\textipa{pt}/, /\textipa{mp}/), 
with each cluster containing a roughly balanced number of reduced and canonical tokens.

To mitigate lexical bias from high-frequency words within clusters (e.g., \textit{just} dominating /\textipa{st}/, \textit{and} dominating /\textipa{nd}/), we downsampled dominant word types to a maximum of 400 tokens per word (200 reduced + 200 canonical). This ensured probing performance reflects cluster-level phonological contrasts rather than word-specific lexical memorization by the classifiers. The final dataset comprised 6,760 tokens (3,409 canonical, 3,351 reduced) across the 7 cluster types as displayed in Table~\ref{tab:dataset-summary}. Token selection was additionally distributed across speakers to improve generalizability across speakers.
%the influence of speaker-specific productions.
%and improve generalizability. 

\begin{table}[h]
\centering
\caption{AAE CCR balanced dataset summary}
\label{tab:dataset-summary}
\begin{tabular}{lccc}
\toprule
\textbf{Cluster Type} & \textbf{Canonical} & \textbf{Reduced} & \textbf{Total} \\
\midrule
\textipa{/pt/}    & 43   & 43   & 86   \\
\textipa{/mp/}    & 45   & 45   & 90   \\
\textipa{/sk/}    & 67   & 67   & 134  \\
\textipa{/ft/}    & 103  & 103  & 206  \\
\textipa{/nt/}    & 917  & 919  & 1,836\\
\textipa{/nd/}    & 990  & 965  & 1,955\\
\textipa{/st/}    & 1,244& 1,209& 2,453\\
\midrule
\textbf{Total}  & \textbf{3,409}& \textbf{3,351}& \textbf{6,760} \\
\bottomrule
\end{tabular}
\end{table}

\subsection{Models and representations}
\subsubsection{Transformer-based models}

This study probes representations from two prominent speech models: \textit{wav2vec2-base} \cite{baevski2020wav2vec2} and \textit{Whisper-small} \cite{radford2023robust}. Both models employ a convolutional feature extractor followed by 12 transformer layers with 768 embedding dimensions, enabling direct layer-wise comparison of how they encode phonological phenomena such as CCR.

\textit{wav2vec2-base} is pretrained in a self-supervised manner on 960 hours of English LibriSpeech audio using a contrastive objective over quantized latent targets, learning representations from raw acoustics without any textual supervision. In contrast, \textit{Whisper-small} is trained in a supervised multilingual multitask setting on approximately 680,000 hours of audio-text data (approximately 65\% English ASR) in a sequence-to-sequence framework, exposing it to large amounts of transcript-aligned and cross-lingual supervision. We probe only Whisper's encoder because it extracts contextualized frame representations from raw spectrograms, while the decoder handles text generation \cite{radford2023robust}. We use the \emph{base} (wav2vec 2.0) and \emph{small} (Whisper) variants because they have the same number of Transformer encoder layers and are both used in their pretrained form without any task-specific fine-tuning. All models remain frozen during probing, with only linear classifiers trained on extracted hidden states. This setup attributes differences in probe performance to pretraining strategies rather than model depth, adaptation, or downstream fine-tuning.

% By comparing the two, we test whether emergent phonological knowledge, such as sensitivity to or restoration of canonical forms in dialectal reductions like CCR, arises primarily from acoustic invariances learned in a self-supervised manner (wav2vec 2.0) or benefits from lexical and transcript-level priors introduced through large-scale supervised multitask training (Whisper). This contrast matters especially for dialects like AAE, where phonological reductions follow systematic rules but produce forms that deviate from standard orthographic representations.

\subsubsection{Representation extraction}

Hidden states were extracted from all 12 transformer layers of both \textit{wav2vec2-base} and \textit{Whisper-small} for the entire dataset. For each token, the full utterance containing the target consonant cluster was fed into the frozen encoder. Using MFA timestamps, temporal boundaries of the consonant cluster (C1C2) relative to utterance onset were identified and converted to frame indices based on each model’s native 20 ms frame resolution.

For wav2vec 2.0, we used the official \texttt{facebook/wav2vec2-base} checkpoint. For Whisper, we extracted representations exclusively from the encoder of \texttt{openai/whisper-small}, consistent with prior probing work.

Aligned frames corresponding to the full cluster (CC) and the reduced onset (C1) were mean-pooled to obtain a single 768-dimensional representation per layer per token, following standard practice in speech probing \cite{yang23v_interspeech, mohebbi-etal-2023-homophone}. This converts variable-length frame sequences  (e.g., 5×768 for short clusters) into fixed-length token embeddings (1×768) while preserving phonetic alignment. For each model, this procedure yielded 12 layer-wise datasets of size (6760, 768).

%All models were kept frozen throughout extraction; only downstream linear probes were trained. This pooling-based extraction strategy ensures that phonologically meaningful units are represented by single, comparable embeddings, while remaining agnostic to model-specific front-end processing. Crucially, it allows us to probe how information about CCR is distributed across layers and architectures, despite differences in pretraining objectives and input representations.

\subsection{Probing experiments}

We performed two domain-informed probing tasks on the frozen representations of \textit{wav2vec2-base} and \textit{Whisper-small}: (1) \emph{Experiment 1} as segmental reduction detection, and (2) \emph{Experiment 2} as segmental restoration.

We adopted a relatively simple architecture to reduce the influence of probe complexity, so that performance differences more directly reflect differences in the information encoded in the embeddings rather than the complexity of the probing model. Specifically, we used scikit-learn (v. 1.7.0) \cite{Pedregosa_2011} multi-layer perceptrons (MLPs) comprised of a single hidden layer of 200 ReLU neurons (\texttt{hidden\_layer\_sizes=(200,)}) followed by a single logistic output neuron for binary classification. The models used default hyperparameters except for the expanded hidden layer size, \texttt{max\_iter=1000}, and \texttt{random\_state=42} for reproducibility. No hyperparameter tuning was performed on the probe architecture itself, further ensuring that differences in probe performance reflect differences in the underlying representations rather than probe optimization. 

For all subsequent probes, we used Stratified KFold cross validation (K=4) with speaker-independent splits (no overlap between training and test speakers). Because this constraint limited how tokens could be distributed across folds, test sets comprised approximately 20-30\% of tokens rather than exactly 25\%.

\subsubsection{Experiment 1 (segmental reduction detection)}\label{sec:method:probexp:exp1}

For the first domain-informed probe, we trained layer-wise MLPs to distinguish \textit{reduced} vs. \textit{canonical} CCR realizations across our dataset. The probe was trained and tested on both reduced and canonical tokens. To maximize generalizability while minimizing lexical and cluster-type biases, we evaluated performance under three dataset conditions:

\begin{enumerate}
    \item \textbf{Imbalanced}: From the 6,760 available tokens, we retained 6,698 such that each cluster type was internally balanced (reduced = canonical), while preserving the natural frequency distribution of CCR cluster types. This condition maintains \textit{ecological validity in cluster prevalence} while controlling for reduction frequency within each cluster.
    
    \item \textbf{Balanced}: All seven clusters were further downsampled to match the cluster type with the fewest available tokens (N = 676 total). In this condition, clusters are equal in size and each remains internally balanced (reduced = canonical). This controls for \textit{quantity biases}, where frequent clusters might inflate performance due to model overfit rather than true CCR detection. Whenever possible, tokens were sampled from different words and speakers to enhance generalizability.
    
    \item \textbf{Per-cluster analysis}: Each of the seven cluster types was probed independently using all available tokens for that cluster, with internal balancing maintained. This isolates \textit{cluster-conditioned effects}, as prior AAE research shows that CCR reduction patterns vary by cluster type \cite{Erik_Baily_2015,Bayley_Villarreal_2019}, motivating separate analyses alongside aggregate results. 
\end{enumerate}

This multi-level analysis reveals how the models perform CCR detection across varying data quantities, cluster type compositions, and individual phonological environments.

\begin{figure*}[t]
  \centering
  \includegraphics[width=0.9\textwidth]{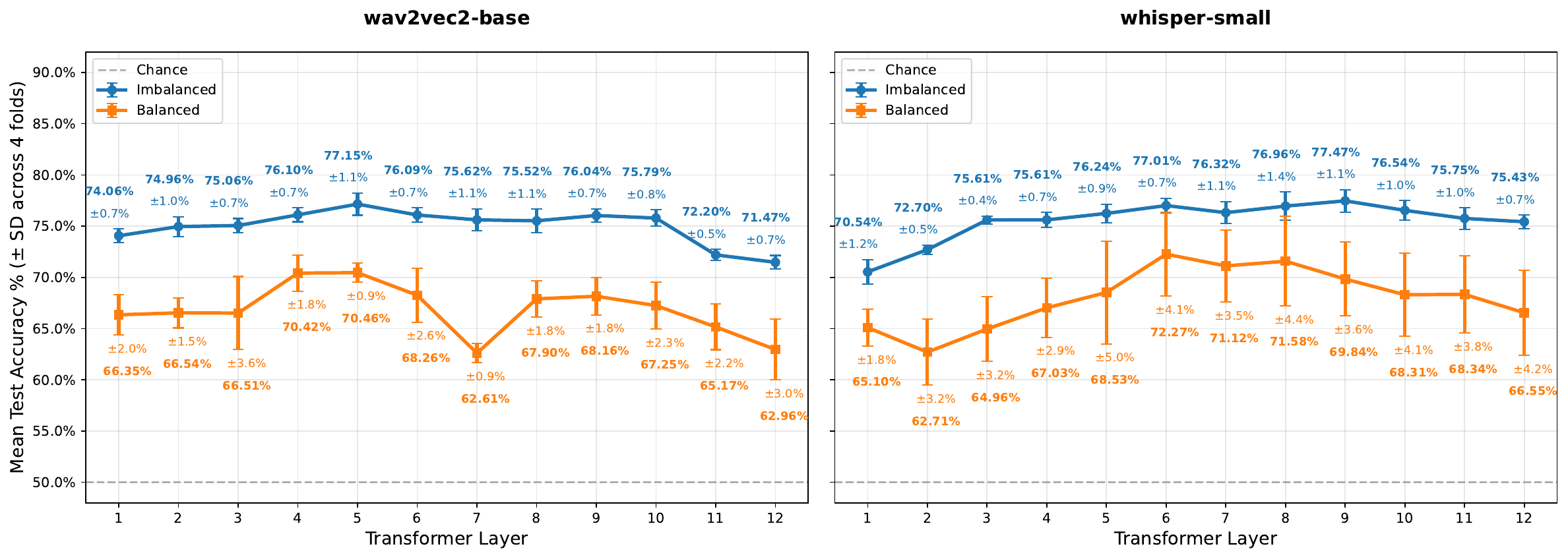}
  \caption{Segmental reduction detection - imbalanced vs. balanced for all cluster types (4-fold CV)}
  \label{fig:binary_classification_balanced_imbalanced}
\end{figure*}

\subsubsection{Experiment 2 (segmental restoration)}\label{sec:method:probexp:exp2}

For the second domain-informed probe, we selected /\textipa{nt}/ and /\textipa{nd}/ tokens from our 7 cluster types because they share the same C1 (nasal /\textipa{n}/) while differing only in C2 (/\textipa{t}/ vs. /\textipa{d}/). This eliminates C1 variation as a confound, ensuring the probe tests pure C2 recovery (coronal stop identity) from reduced nasal-only input (/\textipa{n}/). In addition, both cluster types had sufficient tokens in reduced and canonical forms to support robust probing.

This probe tested whether models' representations of reduced nasal-only segments (/\textipa{n}/ from reduced /\textipa{nt, nd}/ clusters) contain sufficient segmental restoration cues to recover canonical cluster identity (/\textipa{nt}/ vs. /\textipa{nd}/). To mitigate the influence of highly frequent words in these two cluster types, we further limited each word to maximally 100 tokens (50 reduced, 50 canonical), ensuring a roughly balanced contribution of frequent and less frequent words. We conducted three analyses as follows:

\begin{enumerate}
    \item \textbf{Reduced-only train}: Train the probe on embeddings of C1 (the nasal) extracted from reduced /\textipa{nt, nd}/ tokens and test it on embeddings from canonical /\textipa{nt, nd}/ forms. The goal is to determine whether reduced tokens retain cluster-specific information. In other words, we test whether a reduced nasal from /\textipa{nt}/ correctly predicts canonical /\textipa{nt}/ (and a reduced nasal from /\textipa{nd}/ predicts canonical /\textipa{nd}/), rather than collapsing across categories.

    \item \textbf{Canonical-only train}: Train and test on CC embeddings from canonical /\textipa{nt}, \textipa{nd}/ tokens. This is expected to reveal better performance when \textipa{/t, d/} articulation is fully present.

    \item \textbf{C1-only train}: Train on C1 (nasal) embeddings from canonical /\textipa{nt}, \textipa{nd}/ tokens, test on full CC embeddings from canonical /\textipa{nt, nd}/. This probe tests if nasal alone in the train set provides \textipa{/t, d/} cues when full cluster is available at test time.
\end{enumerate}

Table~\ref{tab:cv_splits_compact} shows the train and test set sizes for each probe across the four cross-validation folds. Test sizes are reduced to ensure speaker-independent splits, resulting in fewer tokens than the total available per category.

\begin{table}[h!]
\footnotesize
\centering
\setlength{\tabcolsep}{4pt}   % default is ~6pt → tighter columns
\caption{Train and test sizes per 4-fold CV split}
\label{tab:cv_splits_compact}
\begin{tabular}{l@{\hspace{6pt}}c@{\hspace{6pt}}c}
\toprule
\textbf{Probe/Cluster} & \textbf{Train} & \textbf{Test} \\
\midrule
Imbalanced  & [4933, 5118, 5022, 5021] & [1765, 1580, 1676, 1677] \\
Balanced    & [526, 469, 505, 528]     & [150, 207, 171, 148] \\
\midrule
Reduced-Only   & [1271, 1265, 1263, 1274] & [489, 483, 457, 478] \\
Canonical-Only & [1286, 1262, 1293, 1294] & [477, 477, 477, 476] \\
C1-Only        & [1286, 1262, 1293, 1294] & [477, 477, 477, 476] \\
\midrule
\textipa{/st/} & [1825, 1840, 1927, 1662] & [593, 578, 491, 756] \\
\textipa{/nd/} & [1553, 1489, 1428, 1320] & [377, 441, 502, 610] \\
\textipa{/nt/} & [1407, 1471, 1357, 1267] & [427, 363, 477, 567] \\
\textipa{/ft/} & [163, 164, 148, 143] & [43, 42, 58, 63] \\
\textipa{/sk/} & [94, 100, 110, 98]   & [40, 34, 24, 36] \\
\textipa{/mp/} & [79, 58, 67, 66]     & [11, 32, 23, 24] \\
\textipa{/pt/} & [68, 62, 61, 67]     & [18, 24, 25, 19] \\
\bottomrule
\end{tabular}
\end{table}

\subsubsection{Coarticulatory encoding probe}\label{sec:method:probexp:coarticulatoryprob}

Following the main probing experiments, we conducted an additional supplementary analysis to examine whether the models' representations encode coarticulatory information between C1 and C2. This investigation was motivated by the possibility that the relatively multifaceted layer-wise patterns observed in reduction detection (Section~\ref{sec:results:exp1:imbalancevsbalanced}) might reflect coarticulatory cues preserved in C1, rather than a simple segmental deletion distinction between reduced and canonical forms. We used the same speaker-independent 4-fold CV splits as in the imbalanced condition of the segmental reduction detection probe (Table~\ref{tab:cv_splits_compact}).

To test this hypothesis, we adapted the gating paradigm from psycholinguistics \cite{grosjean1980}, in which listeners hear progressively longer portions of a word until identification becomes possible. Here, we operationalized this by extracting C1 frames in four incremental portions (25\%, 50\%, 75\%, and 100\%) based on forced alignment timestamps, then mean-pooling each portion to obtain fixed-dimensional representations. For each portion, we trained MLP classifiers (identical architecture mentioned earlier) to predict reduced versus canonical CCR forms without including any C2 frames. This design allows us to assess how much information about the reduction contrast is recoverable from C1 alone, and whether coarticulatory information increases as more of C1 becomes available.

\begin{figure*}[t]
  \centering
  \includegraphics[width=0.9\textwidth]{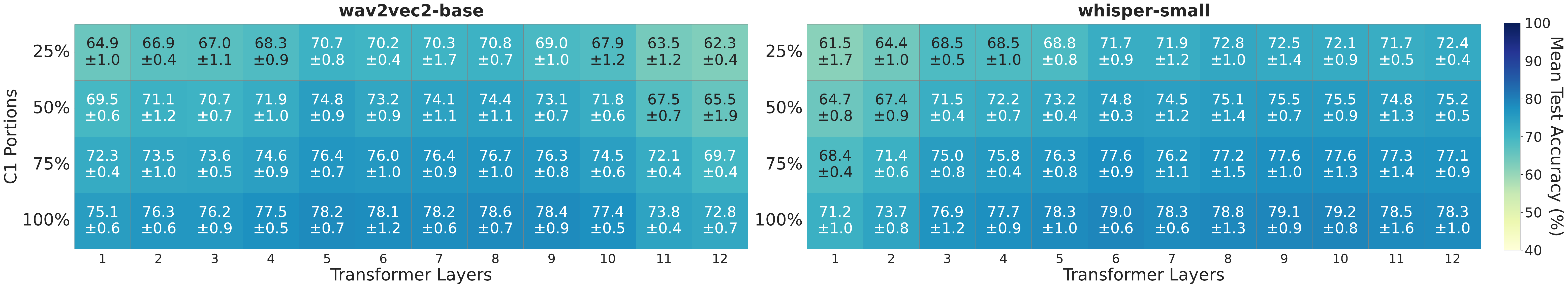}
  \caption{Coarticulatory encoding - C1 portion accuracy per layer across both models (4-fold CV)}
  \label{fig:c1_portions_side_by_side_both_models}
\end{figure*}

\section{Results}
\subsection{Experiment 1}\label{sec:results:exp1}
\subsubsection{Imbalanced vs. balanced}\label{sec:results:exp1:imbalancevsbalanced}
Figure~\ref{fig:binary_classification_balanced_imbalanced} illustrates the layer-wise distribution of information in both wav2vec 2.0 and Whisper representations for the imbalanced and balanced datasets described in Section~\ref{sec:method:probexp:exp1}. Both wav2vec2-base and Whisper-small clearly distinguish reduced from canonical tokens, with accuracies consistently above chance (50\%). This suggests that both models exhibit similar sensitivity to this phenomenon. For the imbalanced dataset, wav2vec2-base accuracy gradually increases in the early layers, peaks at layer 5, then slightly declines before rising again at layer 9, followed by a moderate drop after layer 10 toward the final layers. The initial theoretical interpretation is that this task captures a phonetic distinction, differentiating tokens with and without the final stop. This view aligns with Pasad et al. \cite{Pasad2023}’s characterization of phonetic information, which increases from the early layers toward layer 5, exhibits moderate variation, peaks around layer 9, and then sharply declines after layer 10. Our results generally follow this pattern, although in our plot the accuracy peaks more strongly at layer 5 than at layer 9, whereas they report a higher peak at layer 9 than at layer 5. Additionally, the decline after layer 10 in our data is more moderate than what they observed. In contrast, prior analyses of Whisper’s encoder (particularly for larger models) report a rising-then-plateau pattern for phonetic tasks, with performance increasing toward mid-layers and then remaining relatively stable without a sharp decline toward higher layers \cite{Batra_2025, Agaoglu2024, Yue_etal_2026}. Whisper-small, while less studied, appears to follow this rising-then-plateau trajectory in our imbalanced dataset.

For the balanced dataset, wav2vec 2.0 shows a somewhat unstable pattern: accuracy plateaus in the early layers, peaks around layers 4-5, drops sharply toward layer 7, rises again around layers 8-9, and then declines toward the final layers. Likewise, Whisper does not follow the expected pattern, displaying an early drop at the initial layers (layer 2), a peak at layer 6, and failing to maintain a plateau in the upper layers. On top of this, the overall classification accuracy of the MLP probe is substantially lower in the balanced setting, accompanied by higher standard deviations in both models, particularly for Whisper (SD $\approx$ 4\%). We attribute this decrease in performance primarily to the drastic reduction in training data. Whereas the imbalanced dataset contained 6,698 tokens, the balanced version included only 676 tokens, approximately one tenth of the original size, due to downsampling all cluster types to match the least frequent category (around 100 tokens per cluster). This substantial reduction in training data likely limited the probe’s ability to reliably distinguish reduced from canonical forms. Moreover, in natural speech, certain cluster types occur far more frequently than others, and ASR models are trained on similarly skewed distributions. Artificially downsampling all clusters to the least frequent type therefore creates an evaluation setting that is not representative of the distributional properties the models were trained on. For this reason, and given the clear performance degradation under extreme data reduction, we report only the results obtained from the imbalanced dataset in the remainder of the study.

\begin{figure*}[t]
  \centering
  \includegraphics[width=0.9\textwidth]{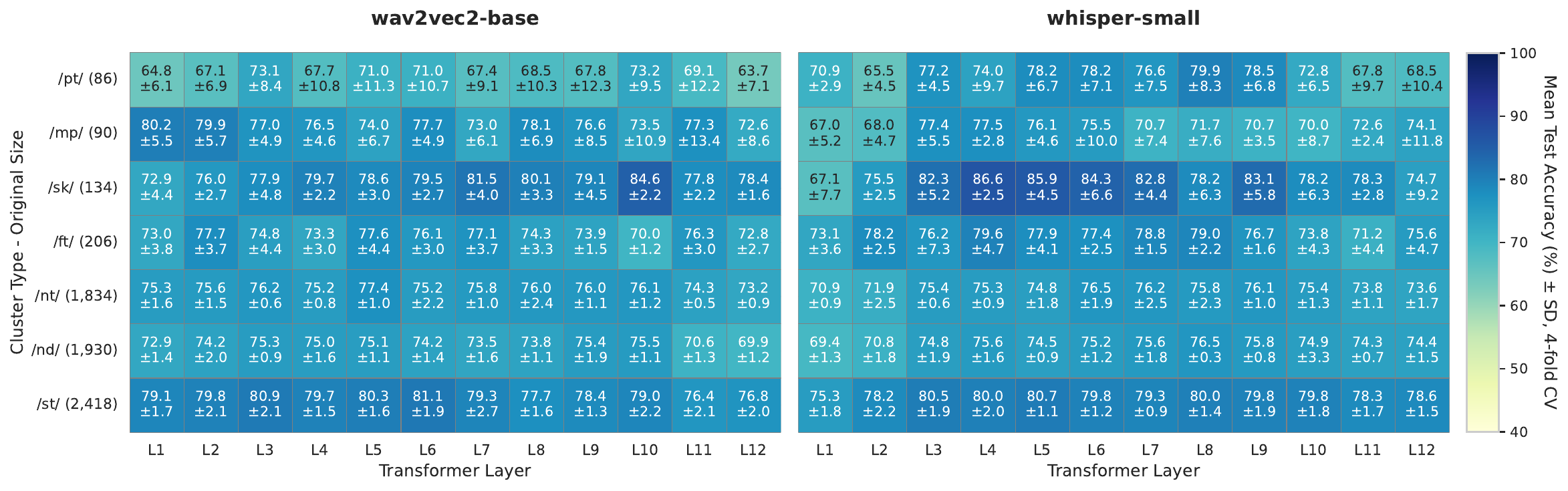}
  \caption{Per-cluster analysis: test accuracy per cluster type (mean (\%) ± SD, 4-fold CV)}
  \label{fig:per_cluster_both_models_heatmap}
\end{figure*}

To further investigate the multifaceted representation of CCR observed in the imbalanced condition, especially for wav2vec 2.0, which cannot be fully attributed to a phonetic distinction (Figure~\ref{fig:binary_classification_balanced_imbalanced}), we implemented the \emph{coarticulatory encoding probe} mentioned in Section~\ref{sec:method:probexp:coarticulatoryprob}. We hypothesized that the reduction detection task may not involve a simple phonetic distinction between reduced and canonical forms, but rather the encoding of coarticulatory or contextual information arising from interaction between the two consonants. This interpretation is consistent with Pasad et al.~\cite{Pasad_2021}, who argue that the layer-wise evolution of wav2vec 2.0 representations follows the linguistic hierarchy of speech understanding, with early layers encoding acoustic features, followed by phonetic information, and higher layers capturing word identity and semantic information. 
Using the gating-style probe described in Section~\ref{sec:method:probexp:coarticulatoryprob}, we found that both wav2vec 2.0 and Whisper achieved high classification accuracy based solely on C1 representations, with performance increasing as larger portions of C1 were included and peaking at 100\%. These results strongly suggest that contextual or coarticulatory information is encoded in the models’ representations, which may help explain the multifaceted representation observed across the hidden layers. Notably, Whisper demonstrates relatively higher performance, as illustrated in Figure ~\ref{fig:c1_portions_side_by_side_both_models}, indicating stronger contextual encoding effects.

% \begin{figure*}[t]
%   \centering
%   \includegraphics[width=\textwidth]{per_cluster_heatmap_both_models_balanced_with_tokens_in_labels.pdf}
%   \caption{Layer-wise Test Accuracy Heatmap per Cluster Type across both models}
%   \label{fig:per_cluster_both_models_heatmap}
% \end{figure*}

\subsubsection{Per cluster analysis}
Figure~\ref{fig:per_cluster_both_models_heatmap} illustrates the layer-wise information encoded in the wav2vec 2.0 and Whisper representations for each cluster type described in Section~\ref{sec:method:probexp:exp1}, with the corresponding dataset sizes indicated on the left. For high-frequency clusters (/\textipa{st}/, /\textipa{nd}/, and /\textipa{nt}/) wav2vec 2.0 again exhibits multiple rises in classification accuracy: an early increase in layers 3-5, followed by a slight decline and plateau, a renewed rise around layers 8-10, and a subsequent decline, mirroring the pattern observed in Figure 1 for the imbalanced condition. Notably, /\textipa{st}/ shows the highest accuracy in the early layers, a pattern discussed from a linguistic perspective in the Discussion section. In contrast, Whisper exhibits a rising-then-plateau pattern that again parallels the pattern observed for the imbalanced condition in Figure~1, with peak performance varying across cluster types and /\textipa{st}/ achieving comparatively higher accuracy. Overall, both models show similar classification accuracy, with wav2vec 2.0 exhibiting slightly higher peaks for two of the three clusters (except \textipa{/nd/}), suggesting comparable effectiveness in detecting CCR with plausible levels of variability (SD).  

The test accuracy becomes even more pronounced for low-frequency clusters (/\textipa{ft}/, /\textipa{sk}/, /\textipa{mp}/, /\textipa{pt}/). Although estimates in these conditions are inherently noisier due to limited data, Whisper shows substantially larger advantages over wav2vec 2.0 across most of these clusters, except for /\textipa{mp}/, where wav2vec 2.0 performs better. However, standard deviations for these low-frequency clusters are notably high in both models, indicating substantial performance variability across cross-validation folds or data subsets. Nonetheless, Whisper's consistently stronger means under data scarcity highlight more robust generalization for CCR detection, likely from its vast weakly supervised training.

% \begin{figure*}[t]
%   \centering
%   \includegraphics[width=\textwidth]{phono_NT_ND_all_variants_side_by_side.pdf}
%   \caption{Phonological Probe (NT vs ND) — All Variants}
%   \label{fig:phono_NT_ND_all_variants_side_by_side}
% \end{figure*}

\begin{figure*}[t]
  \centering
  \includegraphics[width=0.9\textwidth]{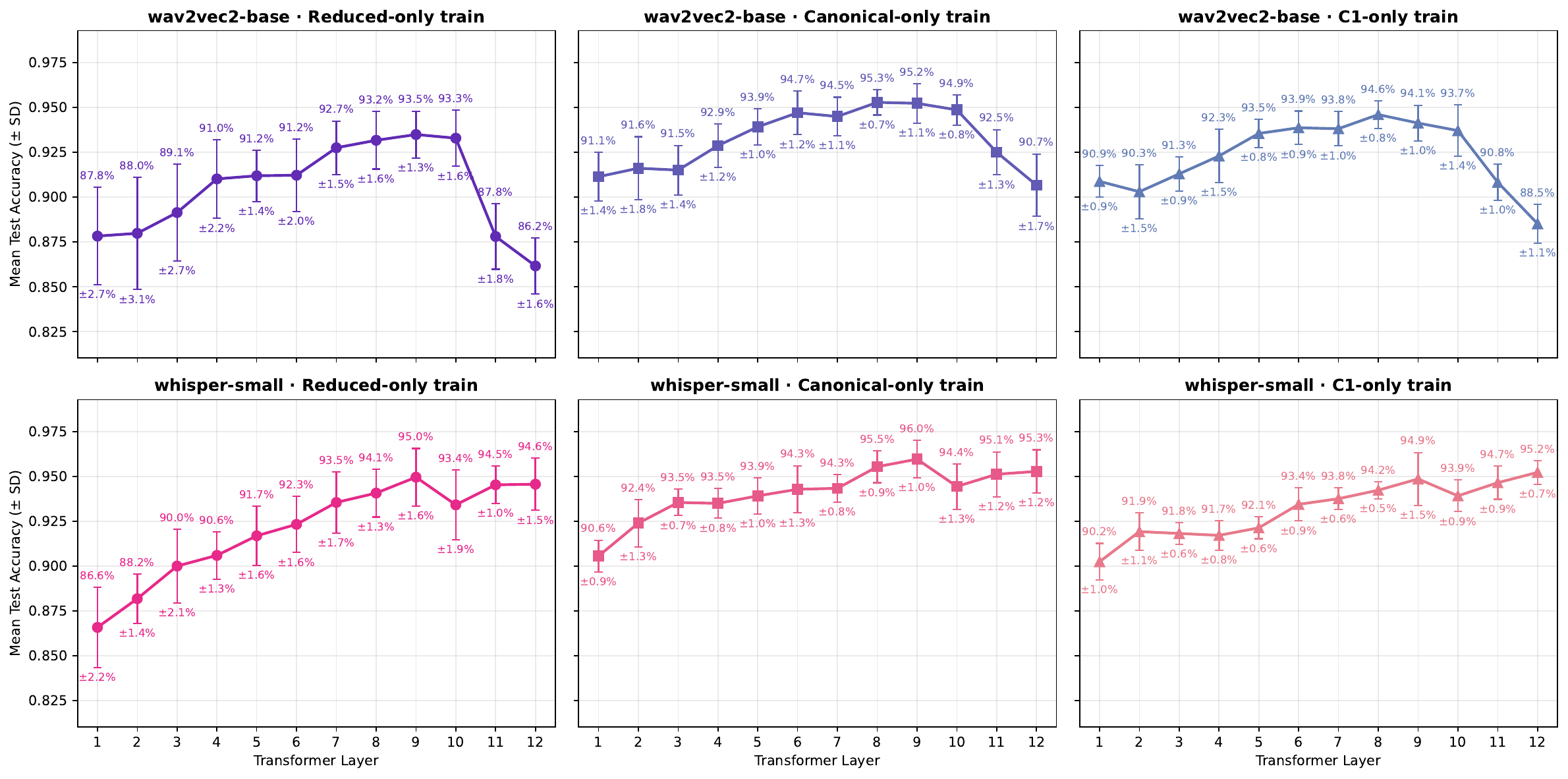}
  \caption{Segmental restoration probe (/\textipa{nt}/ vs /\textipa{nd}/) - all variants (4-fold CV)}
  \label{fig:phono_NT_ND_all_variants_side_by_side}
\end{figure*}

\subsection{Experiment 2}
Figure~\ref{fig:phono_NT_ND_all_variants_side_by_side} summarizes the results of the segmental restoration probe. In contrast to Experiment 1, both models now exhibit smoother, largely single-peaked accuracy curves across layers. For the reduced-only train probe, wav2vec 2.0 shows relatively low accuracy in the earliest layers, followed by a steady increase that peaks around layer 9, before dropping sharply after layer 10. Whisper follows a similar trajectory, but with a steeper rise toward the peak at layer 9, followed by a more moderate decline and a subsequent plateau toward the final layers. The standard deviations are broadly comparable across models, with slightly higher variability for wav2vec 2.0 in the early layers. This probe is theoretically more challenging than the other two shown in the figure, since the reduced forms lack C2, requiring the model to infer the corresponding canonical cluster. Consequently, higher variability across splits is expected, consistent with the observed SD patterns.

Within each model, the three conditions (reduced-only, canonical-only, and C1–only trains) exhibit highly similar layer-wise profiles and comparable peak accuracies (for wav2vec 2.0-base, \( \approx 93\text{-}95\% \); for Whisper-small, \( \approx 94\text{-}96\% \)), indicating that Whisper restores phonological information with slightly higher accuracy overall. The consistent behavior across all three probes for each model suggests that the reduced-only probe performed as expected. As anticipated, CC and C1 embeddings from canonical tokens (canonical-only and C1-only trains) reliably recover their corresponding canonical forms. Importantly, this pattern indicates that reduced embeddings for both cluster types retain information about their canonical counterparts. Both models appear to encode these reductions not merely as surface-level segmental differences, but as systematic phonological variants of the corresponding canonical forms.

% \begin{figure*}[t]
%   \centering
%   \includegraphics[width=0.8\textwidth]{phono_NT_ND_all_variants_side_by_side.pdf}
%   \caption{Phonological Probe (NT vs ND) — All Variants. 
%            Accuracy vs. layer for different C1 portions.}
%   \label{fig:phono_compact}
% \end{figure*}

Crucially, because the coarticulatory effect between C1 and C2 is held constant across cluster types in this probe (both /\textipa{nt}/ and /\textipa{nd}/ share the same C1), the substantially higher accuracies observed here (peaking 93.5\% for wav2vec2-base and 95\% for Whisper-small) indicate that the lower performance in Experiment 1 (70-80\%) (Section~\ref{sec:results:exp1}) should not be interpreted as a failure of the models. Instead, it reflects the fact that CCR tokens retain substantial information about the dropped stop: as shown by both the segmental restoration and the gating-based probes, reduced tokens remain very close to their canonical counterparts in the models’ embedding spaces, so there is no sharp phonetic boundary for a simple MLP to exploit in a binary reduced/canonical classification. In other words, CCR is treated as a case of segmental restoration of an underlying stop, with reduced realizations forming a continuum of gradience rather than a clean-cut contrast between full and reduced pronunciations. This gradient pattern, and its cluster-type-specific manifestations, is discussed in more detail from a linguistic perspective in the Discussion section.

The single-peaked, mid-to-late layer profiles we observe in the segmental restoration probes align with earlier work on phonetic probing of wav2vec 2.0, which reports maximal phonetic category and feature information in intermediate transformer layers \cite{cormac-english-etal-2022-domain, delafuente24_interspeech, kim24l_interspeech}. English et al. \cite{cormac-english-etal-2022-domain} probe layer-wise wav2vec 2.0 embeddings with a simple classifier and show that the transformer’s contextualization enhances phonetic detail, yielding the highest accuracies for phoneme and articulatory feature classification in intermediate layers (layer 9). 

For Whisper, our findings align with recent layer-wise probing studies showing that encoder representations capture phonetic and phonological properties. Classifiers trained on these embeddings reliably detect phonetically deviant speech, typically peaking in mid-to-late encoder layers \cite{Batra_2025, Yue_etal_2026}. Closer to our approach, Gessinger et al. \cite{Shams2025} demonstrate phonemic restoration in wav2vec 2.0 and Whisper using controlled deletions (noise/silent gaps). Like their artificial ``gaps'', our natural CCR deletions (/nt/ → /n/) create analogous disruptions that models successfully restore. Our segmental restoration probe mirrors this pattern: reduced vs. canonical CCR tokens achieve high accuracies in both wav2vec 2.0 and Whisper, encoding cluster reduction as phonologically structured variation rather than mere absence, akin to human listeners achieving \textit{phonological abstraction} without explicit phoneme representations \cite{Scharenborg2013}, where context-specific allophones rather than canonical phonemes serve as prelexical processing units, enabling robust word recognition from gradient input.

\subsection{Control probes}
We verified the reliability of our results for both Experiment 1 and Experiment 2 probes and ruled out potential artifacts due to data leakage or overfitting. To do so, we conducted control experiments on both the imbalanced dataset (Experiment 1) and the reduced-only train probe (Experiment 2). In each case, reduced/canonical labels were randomly shuffled while preserving original train/test splits, speaker independence, and token-cluster distributions. MLP probes trained on these shuffled labels yielded a baseline performance floor of 46-53\%, confirming that the observed accuracies exceed chance and reflect phonetic information encoded in the models’ representations.

\section{Discussion}
In this paper, we examined how layer-wise probing results can inform our understanding of phonological structure, focusing in particular on consonant cluster reduction, coronal stop deletion, and broader patterns in consonant cluster typology. While the Results section emphasized the computational behavior of the models, tracing how representations evolved across layers and experimental conditions, the present discussion reframed these findings through a linguistic lens.

Work on CCR in AAE shows that reduction is not a single, fixed process but a highly conditioned one. Thomas and Bailey \cite{Erik_Baily_2015, wolframThomas} in particular maintain that several structural factors repeatedly emerge as key in CCR production: syllable type, cluster type, and the following environment. In other words, reduction is more common in unstressed than in stressed syllables, more common in homorganic clusters (where C1 and C2 share voicing and place, e.g. \textit{cold}, \textit{fist}) than in non‑homorganic ones (\textit{mint}, \textit{felt}), and more common in monomorphemic clusters (\textit{wound}, \textit{bust}) than in bimorphemic clusters where the final stop is a separate morpheme (\textit{bussed}, \textit{stunned}). The segment that follows the cluster also strongly conditions reduction: across English varieties, deletion is more frequent before consonant‑initial words (\textit{first place}) than before vowel‑initial words (\textit{first area}), and least frequent before a pause, with a further hierarchy among following consonants (most deletion before obstruents, then nasals/liquids, then glides). This following‑segment effect can be captured in terms of sonority: the less sonorous the following sound, the more likely cluster simplification becomes.

Bayley and Villarreal \cite{Bayley_Villarreal_2019} also make this structure especially explicit by distilling Labov’s \cite{Labov_1989} findings into six hierarchically ordered constraints for coronal stop deletion ($>$ denotes more likely to be deleted): % (/\textipa{t}/ or /\textipa{d}/ as C2):
\textbf{1. Stress}: Unstressed $>$ stressed syllables. \textbf{2. Cluster length}: CCC $>$ CC. \textbf{3. Preceding segment}: /\textipa{s}/ $>$ stops$>$ nasals$>$ fricatives$>$ liquids. \textbf{4. Grammatical status}: \textit{n't} $>$ part of stem$>$ derivational suffix$>$ past tense/participle.
\textbf{5. Following segment}: obstruents$>$ liquids$>$ glides$>$ vowels$>$ pauses. \textbf{6. Voicing agreement}: Adjacent segments share voicing (homovoiced) $>$ different voicing.

% \begin{enumerate}
% \item \textbf{Stress}: Unstressed $>$ stressed syllables.
% \item \textbf{Cluster length}: CCC $>$ CC.
% \item \textbf{Preceding segment}: /\textipa{s}/ $>$ stops$>$ nasals$>$ fricatives$>$ liquids.
% \item \textbf{Grammatical status}: \textit{n't} $>$ part of stem$>$ derivational suffix$>$ past tense/participle.
% \item \textbf{Following segment}: obstruents$>$ liquids$>$ glides$>$ vowels$>$ pauses.
% \item \textbf{Voicing agreement}: Adjacent segments share voicing $>$ different voicing
% \end{enumerate}

Within this framework, the ``preceding segment'' constraint corresponds to the C1-C2 interaction in our study: clusters with a highly constrained obstruent C1 (especially /\textipa{s}/) create a particularly favorable environment for C2 deletion, whereas clusters with more sonorous or less constrained C1s (nasals, liquids) resist deletion more strongly. Put differently, the manner and place of C1 modulate how much coarticulatory pressure is placed on C2, and thus how distinct reduced and canonical tokens remain in the signal. This maps directly onto our seven cluster types and provides a linguistically grounded lens for interpreting the cluster‑wise probe patterns we observe.

The per-cluster results in Figure~\ref{fig:per_cluster_both_models_heatmap} align closely with phonetic and phonological accounts of CCR production, revealing principled differences in how wav2vec 2.0-base and Whisper-small encode CCR. Following the C1 constraint hierarchy from Bayley \& Villarreal \cite{Bayley_Villarreal_2019}, our alveolar clusters (/\textipa{nt}/, /\textipa{nd}/, /\textipa{st}/) show robust performance across models (low-70s to low-80s accuracy), with trivial performance gaps. /\textipa{st}/ clusters yield the highest accuracies for both models, exactly as predicted: /\textipa{s}/ + stop provides the strongest CCR context \cite{Wolfram_article, Thomas_2007, Erik_Baily_2015}. /\textipa{nt}/ or /\textipa{nd}/ follow with solid but lower performance, matching their intermediate position in the preceding segment constraint (\textipa{s} $>$ other stops $>$ nasals $>$ other fricatives) from Bayley \& Villarreal \cite{Bayley_Villarreal_2019}. This /\textipa{st}/ $>$ /\textipa{nt, nd}/ gradient precisely mirrors the deletion probability hierarchy, /\textipa{s}/ + stop strongest, nasal-stop intermediate, confirming both models capture  Bayley \& Villarreal\cite{Bayley_Villarreal_2019}'s third constraint (C1 manner effect) reliably in their mid-layer phonetic representations.

In contrast, low-frequency clusters (/\textipa{ft}/, /\textipa{sk}/, /\textipa{pt}/) exhibit a different pattern. These clusters are likewise considered homovoiced (with C1 and C2 sharing the same voicing) and are reported to be more susceptible to reduction \cite{Green_2002}. Moreover, Bayley and Villarreal \cite{Bayley_Villarreal_2019} predict that clusters beginning with /\textipa{s}/ (/\textipa{sk}/) should be most prone to reduction, followed by /\textipa{pt}/ (stop–stop), and then /\textipa{ft}/ (fricative–stop). Interestingly, /\textipa{sk}/ yields the highest accuracy in both wav2vec 2.0 and Whisper, consistent with the expectation that stronger reduction leads to more phonetically distinct reduced vs. canonical classes, which are easier for the MLP probe to separate. Whisper achieves particularly high accuracy for /\textipa{sk}/, suggesting that when reduction is more salient, Whisper captures it more precisely. For /\textipa{pt}/, the expectation is met for Whisper, which attains higher peak accuracy than for /\textipa{ft}/. In contrast, wav2vec 2.0 shows lower performance on /\textipa{pt}/ than on /\textipa{ft}/. This discrepancy is likely attributable to the limited amount of training data available for /\textipa{pt}/ (only 86 tokens), which may disproportionately affect wav2vec 2.0’s probe performance under low-resource conditions. For /\textipa{ft}/, which has comparatively more tokens, both models detect reduction reliably. Overall, Whisper appears to maintain stronger performance than wav2vec 2.0 even for low-resource CCR cluster types, suggesting greater robustness under data scarcity.

/\textipa{mp}/ (90 tokens), classified as non-homorganic by Thomas \& Bailey \cite{Erik_Baily_2015} (voiced /\textipa{m}/ + voiceless /\textipa{p}/), is expected to resist reduction most strongly among all clusters. While wav2vec 2.0 behaves differently, showing an early peak of 80.2\% (layer 1) followed by a steady decline to 72.6\%, Whisper aligns more closely with Thomas and Bailey \cite{Erik_Baily_2015}’s expectations, exhibiting the lowest accuracy among all clusters analyzed so far for this non-homorganic type with low susceptibility to reduction. This suggests that Whisper’s representations may more closely mirror human perceptual patterns in this case. Moreover, the scarcity of training data for this cluster yields high variability in performance (SD $\approx$ 13.4\% for wav2vec 2.0 and SD $\approx$ 11.8\% for Whisper), indicating that the results for /\textipa{mp}/ may be less reliable. This instability likely reflects data sparsity and may also be influenced by factors discussed earlier, such as effects of the following word’s onset, stress patterns, and contextual coarticulation.

Taken together, these results allow for a linguistically grounded interpretation of our probing findings. Overall, the differences observed across clusters do not appear to reflect arbitrary model behavior. Instead, they closely align with well-established phonetic and phonological pressures that shape cluster production and reduction.

\section{Conclusion}

This study set out to determine whether self-supervised and supervised models treat CCR in AAE as a low-level segmental deletion phenomenon, or whether they encode it as structured, gradient phonological variation with preserved cues to underlying forms. Layer-wise probing on 6,760 CORAAL tokens revealed that both wav2vec 2.0-base and Whisper-small encode CCR in a phonologically structured manner. 

In the segmental reduction detection probe, the models reliably distinguished reduced from canonical realizations (accuracies typically 70-80\%). Layer-wise patterns partially aligned with established phonetic hierarchies: wav2vec 2.0 showed multiple rises (strong peak around layer 5, renewed increase near layer 9) with a moderate late decline, broadly consistent with the literature but with a sharper mid-layer emphasis and gentler late drop than previously reported. Our gating paradigm analysis of C1 portion embeddings revealed these differences arise from strong coarticulatory effects in CCR, where preceding segments preserve cues to missing stops. Whisper-small exhibited a rising-then-plateau pattern, closely mirroring patterns in phonetic probing of supervised encoders. 

Cluster types were also treated differently by the models due to various interacting effects, one of which being the preceding segment (C1). This highly affects the way CCR takes place and the degree of reduction that differs from one cluster type to another, making the reduction detection harder than a simple complete segmental deletion but rather a gradient phenomenon shaped by cluster-specific phonological constraints.  

The segmental restoration probe provided decisive evidence of phonological encoding: reduced tokens retained robust cues to underlying stop identity (peak accuracies 93-96\%), with single-peaked mid-to-late layer profiles that again resembled intermediate-layer optima for segmental feature recovery in prior work. These results indicate that both self-supervised and supervised models do not normalize CCR as complete deletion but represent it as systematic, contextually licensed phonological variation, with reduced realizations remaining close to their canonical counterparts in the embedding space.

\section{Limitations and future work}
One limitation of this study is that we evaluated only two models, wav2vec 2.0 and Whisper. As a result, it remains unclear to what extent our findings generalize to other speech representation models. In addition, certain CCR cluster types (e.g., /\textipa{mp}/, /\textipa{pt}/) were underrepresented in the data due to their lower frequency in naturally occurring AAE speech. This imbalance may have decreased the robustness of our analyses for these rarer and often more complex clusters, while potentially inflating performance for more frequent clusters such as /\textipa{nt}/ and /\textipa{nd}/. Furthermore, our analysis was restricted to two-consonant clusters in short, monomorphemic words. We did not examine three-consonant clusters (e.g., /\textipa{skt}/, /\textipa{sts}/), which are known to undergo reduction more frequently and may reveal stronger differences in model behavior. Similarly, we excluded bimorphemic forms, which could offer additional insight into how morphological structure interacts with CCR.
Future research should extend this investigation to additional speech models, such as HuBERT \cite{hubert}, WavLM \cite{wavlm}, and MMS \cite{MMS}, and compare base models with larger or fine-tuned variants (e.g., wav2vec2-large or Whisper-large). This would help determine whether the patterns observed here are architecture-specific or more general across model families and scales.

\section{Acknowledgments}
This work is part of HM's PhD research. The authors would like to thank Erfan Amirzadeh Shams for his helpful contribution in refining the research concept and for his technical guidance on the implementation. The authors also thank the anonymous reviewers for their insightful comments and constructive feedback, which helped improve the quality of this work.

\section{Generative AI Use Disclosure}
No Generative AI tools were used to produce the content or results of this paper. Perplexity.ai and Grok were used for English grammar checking, improving sentence readability, suggesting relevant literature, and code refining or debugging.

\bibliographystyle{IEEEtran}
\bibliography{mybib}

@article{koenecke2020racial,
author = {Allison Koenecke  and Andrew Nam  and Emily Lake  and Joe Nudell  and Minnie Quartey  and Zion Mengesha  and Connor Toups  and John R. Rickford  and Dan Jurafsky  and Sharad Goel },
title = {Racial disparities in automated speech recognition},
journal = {Proceedings of the National Academy of Sciences},
volume = {117},
number = {14},
pages = {7684-7689},
year = {2020},
doi = {10.1073/pnas.1915768117},
URL = {https://doi.org/10.1073/pnas.1915768117},
eprint = {https://www.pnas.org/doi/pdf/10.1073/pnas.1915768117}}

@inproceedings{martin20_interspeech,
  title     = {{Understanding Racial Disparities in Automatic Speech Recognition: The Case of Habitual “be”}},
  author    = {Joshua L. Martin and Kevin Tang},
  year      = {2020},
  booktitle = {{Interspeech 2020}},
  pages     = {626--630},
  doi       = {10.21437/Interspeech.2020-2893},
  issn      = {2958-1796},
  URL = {https://doi.org/10.21437/Interspeech.2020-2893}
}

@article{wassink2022uneven,
title = {Uneven success: automatic speech recognition and ethnicity-related dialects},
journal = {Speech Communication},
volume = {140},
pages = {50-70},
year = {2022},
issn = {0167-6393},
doi = {https://doi.org/10.1016/j.specom.2022.03.009},
url = {https://doi.org/10.1016/j.specom.2022.03.009},
author = {Alicia Beckford Wassink and Cady Gansen and Isabel Bartholomew}
}

@inproceedings{mojarad25_interspeech,
  title     = {{Automatic Speech Recognition of African American English: Lexical and Contextual Effects}},
  author    = {Hamid Mojarad and Kevin Tang},
  year      = {2025},
  booktitle = {{Interspeech 2025}},
  pages     = {3883--3887},
  doi       = {10.21437/Interspeech.2025-1511},
  issn      = {2958-1796},
  URL = {https://doi.org/10.21437/Interspeech.2025-1511}
}

@inproceedings{baevski2020wav2vec2,
author = {Baevski, Alexei and Zhou, Henry and Mohamed, Abdelrahman and Auli, Michael},
title = {wav2vec 2.0: a framework for self-supervised learning of speech representations},
year = {2020},
isbn = {9781713829546},
publisher = {Curran Associates Inc.},
address = {Red Hook, NY, USA},
booktitle = {Proceedings of the 34th International Conference on Neural Information Processing Systems},
articleno = {1044},
numpages = {12},
location = {Vancouver, BC, Canada},
series = {NIPS '20},
URL = {https://dl.acm.org/doi/abs/10.5555/3495724.3496768}
}

@inproceedings{radford2023robust,
author = {Radford, Alec and Kim, Jong Wook and Xu, Tao and Brockman, Greg and McLeavey, Christine and Sutskever, Ilya},
title = {Robust speech recognition via large-scale weak supervision},
year = {2023},
publisher = {JMLR.org},
booktitle = {Proceedings of the 40th International Conference on Machine Learning},
articleno = {1182},
numpages = {27},
location = {Honolulu, Hawaii, USA},
series = {ICML'23},
URL = {https://dl.acm.org/doi/10.5555/3618408.3619590}
}

@book{Green_2002, place={Cambridge}, title={{African American English}: A Linguistic Introduction}, publisher={Cambridge University Press}, author={Green, Lisa J.}, year={2002}}

@ARTICLE{Mengesha_2021,
    
AUTHOR={Mengesha, Zion  and Heldreth, Courtney  and Lahav, Michal  and Sublewski, Juliana  and Tuennerman, Elyse },
           
TITLE={{I} don’t Think These Devices are Very Culturally Sensitive.-Impact of Automated Speech Recognition Errors on {African} {Americans}},
          
JOURNAL={Frontiers in Artificial Intelligence},
          
VOLUME={Volume 4 - 2021},
  
YEAR={2021},
  
URL={https://doi.org/10.3389/frai.2021.725911},
  
DOI={10.3389/frai.2021.725911},
  
ISSN={2624-8212}}

@inproceedings{a-shams-etal-2024-uncovering,
    title = "Uncovering Syllable Constituents in the Self-Attention-Based Speech Representations of Whisper",
    author = "A Shams, Erfan  and
      Gessinger, Iona  and
      Carson-Berndsen, Julie",
    editor = "Belinkov, Yonatan  and
      Kim, Najoung  and
      Jumelet, Jaap  and
      Mohebbi, Hosein  and
      Mueller, Aaron  and
      Chen, Hanjie",
    booktitle = "Proceedings of the 7th BlackboxNLP Workshop: Analyzing and Interpreting Neural Networks for NLP",
    month = nov,
    year = "2024",
    address = "Miami, Florida, US",
    publisher = "Association for Computational Linguistics",
    url = "https://doi.org/10.18653/v1/2024.blackboxnlp-1.16",
    doi = "10.18653/v1/2024.blackboxnlp-1.16",
    pages = "238--247"
}

@incollection{Erik_Baily_2015,
    author = {Thomas, Erik R. and Bailey, Guy},
    isbn = {9780199795390},
    title = {Segmental Phonology of {African} {American} {English}},
    booktitle = {The Oxford Handbook of African American Language},
    publisher = {Oxford University Press},
    year = {2015},
    month = {07},
    doi = {10.1093/oxfordhb/9780199795390.013.13},
    url = {https://doi.org/10.1093/oxfordhb/9780199795390.013.13},
    eprint = {https://academic.oup.com/book/0/chapter/212014935/chapter-ag-pdf/44596684/book_28056_section_212014935.ag.pdf},
}

@inproceedings{parra-2025-interpretable,
    title = "Interpretable Sparse Features for Probing Self-Supervised Speech Models",
    author = "Parra, I{\~n}igo",
    editor = "T.y.s.s, Santosh  and
      Shimizu, Shuichiro  and
      Gong, Yifan",
    booktitle = "The 14th International Joint Conference on Natural Language Processing and The 4th Conference of the Asia-Pacific Chapter of the Association for Computational Linguistics",
    month = dec,
    year = "2025",
    address = "Mumbai, India",
    publisher = "Association for Computational Linguistics",
    url = "https://aclanthology.org/2025.ijcnlp-srw.1/",
    pages = "1--9",
    ISBN = "979-8-89176-304-3"
}

@INPROCEEDINGS{Pasad_2021,
  author={Pasad, Ankita and Chou, Ju-Chieh and Livescu, Karen},
  booktitle={2021 IEEE Automatic Speech Recognition and Understanding Workshop (ASRU)}, 
  title={Layer-Wise Analysis of a Self-Supervised Speech Representation Model}, 
  year={2021},
  volume={},
  number={},
  pages={914-921},
  doi={10.1109/ASRU51503.2021.9688093},
  URL = {https://doi.org/10.1109/ASRU51503.2021.9688093}}

@article{Martin2023,
    author = {Martin, Joshua L and Wright, Kelly Elizabeth},
    title = {Bias in Automatic Speech Recognition: The Case of {African} {American} {Language}},
    journal = {Applied Linguistics},
    volume = {44},
    number = {4},
    pages = {613-630},
    year = {2022},
    month = {12},
    issn = {0142-6001},
    doi = {10.1093/applin/amac066},
    url = {https://doi.org/10.1093/applin/amac066},
    eprint = {https://academic.oup.com/applij/article-pdf/44/4/613/51053335/amac066.pdf},
}

@article{Shams2025,
title = {Under the hood: phonemic restoration in transformer-based automatic speech recognition},
journal = {Computer Speech \& Language},
volume = {96},
pages = {101893},
year = {2026},
issn = {0885-2308},
doi = {https://doi.org/10.1016/j.csl.2025.101893},
url = {https://doi.org/10.1016/j.csl.2025.101893},
author = {Iona Gessinger and Erfan A. Shams and Julie Carson-Berndsen}
}

@inproceedings{cormac-english-etal-2022-domain,
    title = "Domain-Informed Probing of wav2vec 2.0 Embeddings for Phonetic Features",
    author = "Cormac English, Patrick  and
      Kelleher, John D.  and
      Carson-Berndsen, Julie",
    editor = "Nicolai, Garrett  and
      Chodroff, Eleanor",
    booktitle = "Proceedings of the 19th SIGMORPHON Workshop on Computational Research in Phonetics, Phonology, and Morphology",
    month = jul,
    year = "2022",
    address = "Seattle, Washington",
    publisher = "Association for Computational Linguistics",
    url = "https://doi.org/10.18653/v1/2022.sigmorphon-1.9",
    doi = "10.18653/v1/2022.sigmorphon-1.9",
    pages = "83--91"
}

@ARTICLE{Kendall_ing_2021,

AUTHOR={Kendall, Tyler  and Vaughn, Charlotte  and Farrington, Charlie  and Gunter, Kaylynn  and McLean, Jaidan  and Tacata, Chloe  and Arnson, Shelby },

TITLE={Considering Performance in the Automated and Manual Coding of Sociolinguistic Variables: Lessons From Variable {(ING)}},

JOURNAL={Frontiers in Artificial Intelligence},

VOLUME={4},

YEAR={2021},


DOI={10.3389/frai.2021.648543},
URL = {https://doi.org/10.3389/frai.2021.648543},
ISSN={2624-8212}}

@inproceedings{mcauliffe17_interspeech,
  author={McAuliffe, Michael and Socolof, Michaela and Mihuc, Sarah and Wagner, Michael and Sonderegger, Morgan},
  title={{Montreal Forced Aligner: Trainable Text-Speech Alignment Using Kaldi}},
  year=2017,
  booktitle={Proc. Interspeech 2017},
  pages={498--502},
  doi={10.21437/Interspeech.2017-1386},
  URL = {https://doi.org/10.21437/Interspeech.2017-1386}
}

@misc{farrington_corpus_2021,
	title = {The {Corpus} of {Regional} {African} {American} {Language}},
	volume = {Version 2023.06},
	url = {https://oraal.uoregon.edu/coraal},
	doi = {10.7264/1ad5-6t35},
	urldate = {2023-10-01},
	author = {Kendall, Tyler and Farrington, Charlie},
	year = {2023},
	note = {Publisher: The Online Resources for African American Language Project},
    location = {Eugene, OR}
}

@inproceedings{yang23v_interspeech,
  title     = {{What Can an Accent Identifier Learn? Probing Phonetic and Prosodic Information in a Wav2vec2-based Accent Identification Model}},
  author    = {Mu Yang and Ram C. M. C. Shekar and Okim Kang and John H. L. Hansen},
  year      = {2023},
  booktitle = {{Interspeech 2023}},
  pages     = {1923--1927},
  doi       = {10.21437/Interspeech.2023-2254},
  issn      = {2958-1796},
  URL = {https://doi.org/10.21437/Interspeech.2023-2254}
}

@inproceedings{mohebbi-etal-2023-homophone,
    title = "Homophone Disambiguation Reveals Patterns of Context Mixing in Speech Transformers",
    author = "Mohebbi, Hosein  and
      Chrupa{\l}a, Grzegorz  and
      Zuidema, Willem  and
      Alishahi, Afra",
    editor = "Bouamor, Houda  and
      Pino, Juan  and
      Bali, Kalika",
    booktitle = "Proceedings of the 2023 Conference on Empirical Methods in Natural Language Processing",
    month = dec,
    year = "2023",
    address = "Singapore",
    publisher = "Association for Computational Linguistics",
    url = "https://doi.org/10.18653/v1/2023.emnlp-main.513",
    doi = "10.18653/v1/2023.emnlp-main.513",
    pages = "8249--8260"
}

@article{Pedregosa_2011,
  author  = {Fabian Pedregosa and Ga{{\"e}}l Varoquaux and Alexandre Gramfort and Vincent Michel and Bertrand Thirion and Olivier Grisel and Mathieu Blondel and Peter Prettenhofer and Ron Weiss and Vincent Dubourg and Jake Vanderplas and Alexandre Passos and David Cournapeau and Matthieu Brucher and Matthieu Perrot and {{\'E}}douard Duchesnay},
  title   = {Scikit-learn: Machine Learning in Python},
  journal = {Journal of Machine Learning Research},
  year    = {2011},
  volume  = {12},
  number  = {85},
  pages   = {2825--2830},
  url     = {http://jmlr.org/papers/v12/pedregosa11a.html}
}

@inbook{Bayley_Villarreal_2019, place={Cambridge}, series={Studies in English Language}, title={Coronal Stop Deletion in a Rural South Texas Community}, booktitle={Mexican American English: Substrate Influence and the Birth of an Ethnolect}, publisher={Cambridge University Press}, author={Bayley, Robert and Villarreal, Dan}, editor={Thomas, Erik R.Editor}, year={2019}, pages={198–214}, collection={Studies in English Language},
URL = {https://doi.org/10.1017/9781316162316.008}}

@article{Labov_1989, title={The child as linguistic historian}, volume={1}, DOI={10.1017/S0954394500000120}, number={1}, journal={Language Variation and Change}, author={Labov, William}, year={1989}, pages={85–97},
URL = {https://doi.org/10.1017/S0954394500000120}}

@article{Wolfram_article,
author = {Wolfram, Walt},
year = {2003},
month = {06},
pages = {282-316},
title = {Reexamining the Development of {African} {American} {English}: Evidence from Isolated Communities},
volume = {79},
journal = {Language},
doi = {10.1353/lan.2003.0144},
URL = {https://doi.org/10.1353/lan.2003.0144}
}

@article{Thomas_2007,
author = {Thomas, Erik R.},
title = {Phonological and Phonetic Characteristics of {African} {American} {Vernacular} {English}},
journal = {Language and Linguistics Compass},
volume = {1},
number = {5},
pages = {450-475},
doi = {https://doi.org/10.1111/j.1749-818X.2007.00029.x},
url = {https://doi.org/10.1111/j.1749-818X.2007.00029.x},
eprint = {https://compass.onlinelibrary.wiley.com/doi/pdf/10.1111/j.1749-818X.2007.00029.x},
year = {2007}
}

@inproceedings{Batra_2025,
  author={Batra, Ashita and Kar, Brajesh and Das, Pradip K},
  booktitle={2025 33rd European Signal Processing Conference (EUSIPCO)}, 
  title={Exploring {Whisper} Embeddings for Stutter Detection: A Layer-Wise Study}, 
  year={2025},
  volume={},
  number={},
  pages={61-65},
  doi={10.23919/EUSIPCO63237.2025.11226086},
  URL = {https://doi.org/10.23919/EUSIPCO63237.2025.11226086}}

@techreport{Agaoglu2024,
  title = {How Does {OpenAI}'s {Whisper} Interpret Dysarthric Speech?},
  author = {Agaoglu, O. and Yue, Z. and Zhang, Y.},
  year = {2024},
  institution = {TU Delft},
  url = {https://resolver.tudelft.nl/uuid:47837feb-1b2e-4bba-9fad-f92d84024abb},
  note = {Student thesis/project report}
}

@article{grosjean1980,
  author  = {Grosjean, François},
  title   = {Spoken word recognition processes and the gating paradigm},
  journal = {Perception \& Psychophysics},
  year    = {1980},
  volume  = {28},
  number  = {4},
  pages   = {267--283},
  doi     = {10.3758/BF03204386},
  url     = {https://doi.org/10.3758/BF03204386},
  issn    = {1532-5962},
}

@inproceedings{delafuente24_interspeech,
  title     = {{A layer-wise analysis of Mandarin and English suprasegmentals in SSL speech models}},
  author    = {Anton {de la Fuente} and Dan Jurafsky},
  year      = {2024},
  booktitle = {{Interspeech 2024}},
  pages     = {1290--1294},
  doi       = {10.21437/Interspeech.2024-2341},
  issn      = {2958-1796},
  URL = {https://doi.org/10.21437/Interspeech.2024-2341}
}

@misc{Yue_etal_2026,
      title={Probing {Whisper} for Dysarthric Speech in Detection and Assessment}, 
      author={Zhengjun Yue and Devendra Kayande and Zoran Cvetkovic and Erfan Loweimi},
      year={2025},
      eprint={2510.04219},
      archivePrefix={arXiv},
      primaryClass={eess.AS},
      url={https://doi.org/10.48550/arXiv.2510.04219}, 
}

@inproceedings{kim24l_interspeech,
  title     = {{Using wav2vec 2.0 for phonetic classification tasks: methodological aspects}},
  author    = {Lila Kim and Cédric Gendrot},
  year      = {2024},
  booktitle = {{Interspeech 2024}},
  pages     = {1530--1534},
  doi       = {10.21437/Interspeech.2024-1155},
  issn      = {2958-1796},
  URL = {https://doi.org/10.21437/Interspeech.2024-1155}
}

@book{wolframThomas,
  title={The development of African American English},
  author={Wolfram, Walt and Thomas, Erik R},
  year={2002},
  publisher={Blackwell Publishers},
  doi={10.1002/9780470690178},
  isbn={9780631230861},
  URL = {https://doi.org/10.1002/9780470690178}
}

@INPROCEEDINGS{Pasad2023,
  author={Pasad, Ankita and Shi, Bowen and Livescu, Karen},
  booktitle={ICASSP 2023 - 2023 IEEE International Conference on Acoustics, Speech and Signal Processing (ICASSP)}, 
  title={Comparative Layer-Wise Analysis of Self-Supervised Speech Models}, 
  year={2023},
  volume={},
  number={},
  pages={1-5},
  doi={10.1109/ICASSP49357.2023.10096149},
  URL = {https://doi.org/10.1109/ICASSP49357.2023.10096149}}

@inbook{Wolfram_2017,
author = {Wolfram, Walt},
publisher = {John Wiley \& Sons, Ltd},
isbn = {9781405166256},
title = {Dialect in Society},
booktitle = {The Handbook of Sociolinguistics},
chapter = {7},
pages = {107-126},
doi = {https://doi.org/10.1002/9781405166256.ch7},
url = {https://doi.org/10.1002/9781405166256.ch7},
eprint = {https://onlinelibrary.wiley.com/doi/pdf/10.1002/9781405166256.ch7},
year = {2017}
}

@article{Scharenborg2013,
title = {Phonological abstraction without phonemes in speech perception},
journal = {Cognition},
volume = {129},
number = {2},
pages = {356-361},
year = {2013},
issn = {0010-0277},
doi = {https://doi.org/10.1016/j.cognition.2013.07.011},
url = {https://doi.org/10.1016/j.cognition.2013.07.011},
author = {Holger Mitterer and Odette Scharenborg and James M. McQueen}
}

@article{hubert,
  author={Hsu, Wei-Ning and Bolte, Benjamin and Tsai, Yao-Hung Hubert and Lakhotia, Kushal and Salakhutdinov, Ruslan and Mohamed, Abdelrahman},
  journal={IEEE/ACM Transactions on Audio, Speech, and Language Processing}, 
  title={HuBERT: Self-Supervised Speech Representation Learning by Masked Prediction of Hidden Units}, 
  year={2021},
  volume={29},
  number={},
  pages={3451-3460},
  doi={10.1109/TASLP.2021.3122291},
  url = {https://doi.org/10.1109/TASLP.2021.3122291}}

@ARTICLE{wavlm,
  author={Chen, Sanyuan and Wang, Chengyi and Chen, Zhengyang and Wu, Yu and Liu, Shujie and Chen, Zhuo and Li, Jinyu and Kanda, Naoyuki and Yoshioka, Takuya and Xiao, Xiong and Wu, Jian and Zhou, Long and Ren, Shuo and Qian, Yanmin and Qian, Yao and Wu, Jian and Zeng, Michael and Yu, Xiangzhan and Wei, Furu},
  journal={IEEE Journal of Selected Topics in Signal Processing}, 
  title={WavLM: Large-Scale Self-Supervised Pre-Training for Full Stack Speech Processing}, 
  year={2022},
  volume={16},
  number={6},
  pages={1505-1518},
  doi={10.1109/JSTSP.2022.3188113},
  url = {https://doi.org/10.1109/JSTSP.2022.3188113}}

@article{MMS,
  author  = {Vineel Pratap and Andros Tjandra and Bowen Shi and Paden Tomasello and Arun Babu and Sayani Kundu and Ali Elkahky and Zhaoheng Ni and Apoorv Vyas and Maryam Fazel-Zarandi and Alexei Baevski and Yossi Adi and Xiaohui Zhang and Wei-Ning Hsu and Alexis Conneau and Michael Auli},
  title   = {Scaling Speech Technology to 1,000+ Languages},
  journal = {Journal of Machine Learning Research},
  year    = {2024},
  volume  = {25},
  number  = {97},
  pages   = {1--52},
  url     = {http://jmlr.org/papers/v25/23-1318.html}
}

@book{Labov_1972,
  author    = {Labov, William},
  title     = {Sociolinguistic Patterns},
  year      = {1972},
  publisher = {University of Pennsylvania Press},
  address   = {Philadelphia}
}

@book{Schreier_2005,
  author    = {Schreier, Daniel},
  title     = {Consonant Change in {English} Worldwide: Synchrony Meets Diachrony},
  year      = {2005},
  publisher = {Palgrave Macmillan},
  address   = {New York}
}

@article{Guy_1991,
  author  = {Guy, Gregory R.},
  title   = {Explanation in variable phonology: An exponential model of morphological constraints},
  journal = {Language Variation and Change},
  year    = {1991},
  volume  = {3},
  number  = {1},
  pages   = {1--22},
  doi = {10.1017/S0954394500000429},
  URL = {https://doi.org/10.1017/S0954394500000429}
}

\end{document}